%% file: sn-article.tex
\documentclass[sn-mathphys,Numbered]{sn-jnl}% Math and Physical Sciences Reference Style
\usepackage{graphicx}%
\usepackage{multirow}%
\usepackage{amsmath,amssymb,amsfonts}%
\usepackage{amsthm}%
\usepackage{mathrsfs}%
\usepackage[title]{appendix}%
\usepackage{textcomp}%
\usepackage{manyfoot}%
\usepackage{booktabs}%
\usepackage{algorithm}%
\usepackage{algorithmicx}%
\usepackage{algpseudocode}%
\usepackage{listings}%
\graphicspath{ {./images/} }

\usepackage{nameref,hyperref}
\usepackage{longtable}
\usepackage{colortbl}
\usepackage[right]{lineno}
\usepackage[table]{xcolor}

\usepackage{booktabs}
\usepackage{array}

\theoremstyle{thmstyleone}%
\theoremstyle{thmstyletwo}%

\theoremstyle{thmstylethree}%

\begin{document}

\title[Article Title]{Social Robots for Sleep Health: A Scoping Review}


\author*[1]{\fnm{Victor} \sur{Antony}}\email{vantony1@jhu.edu}
\equalcont{These authors contributed equally to this work.}

\author[2]{\fnm{Mengchi} \sur{Li}}\email{mli121@jhu.edu}
\equalcont{These authors contributed equally to this work.}

\author[2,3]{\fnm{Shu-Han} \sur{Lin}}\email{slin98@jhu.edu}

\author[2]{\fnm{Junxin} \sur{Li}}\email{junxin.li@jhu.edu}

\author[1]{\fnm{Chien-Ming} \sur{Huang}}\email{chienming.huang@jhu.edu}

\affil*[1]{\orgdiv{Department of Computer Science}, \orgname{Johns Hopkins University}, \orgaddress{\street{3400 N Charles Street}, \city{Baltimore}, \postcode{21218}, \state{MD}, \country{USA}}}

\affil[2]{\orgdiv{School of Nursing}, \orgname{Johns Hopkins University}, \orgaddress{\street{525 N Wolfe St}, \city{Baltimore}, \postcode{21205}, \state{MD}, \country{USA}}}

\affil[3]{\orgdiv{Department of Nursing}, \orgname{National Taiwan University Hospital}, \orgaddress{\street{No.7, Chung Shan S. Rd.}, \city{Zhongzheng}, \postcode{100}, \state{Taipei}, \country{Taiwan}}}


%\abstract{The abstract serves both as a general introduction to the topic and as a brief, non-technical summary of the main results and their implications. Authors are advised to check the author instructions for the journal they are submitting to for word limits and if structural elements like subheadings, citations, or equations are permitted.}

%%================================%%
%% Sample for structured abstract %%
%%================================%%

\abstract{Poor sleep health is an increasingly concerning public healthcare crisis, especially when coupled with a dwindling number of health professionals qualified to combat it. However, there is a growing body of scientific literature on the use of digital technologies in supporting and sustaining individuals' healthy sleep habits. Social robots are a relatively recent technology that has been used to facilitate health care interventions and may have potential in improving sleep health outcomes, as well. Social robots' unique characteristics---such as anthropomorphic physical embodiment or effective communication methods---help to engage users and motivate them to comply with specific interventions, thus improving the interventions' outcomes.
This scoping review aims to evaluate current scientific evidence for employing social robots in sleep health interventions, identify critical research gaps, and suggest future directions for developing and using social robots to improve people's sleep health. Our analysis of the reviewed studies found them limited due to a singular focus on the older adult population, use of small sample sizes, limited intervention durations, and other compounding factors. Nevertheless, the reviewed studies reported several positive outcomes, highlighting the potential social robots hold in this field. Although our review found limited clinical evidence for the efficacy of social robots as purveyors of sleep health interventions, it did elucidate the potential for a successful future in this domain if current limitations are addressed and more research is conducted.}
% We hope to see social robots being increasingly leveraged to improve sleep health therapies and aid in countering the impending sleep health crisis.

% This scoping review was conducted across the following academic databases: IEEE Xplore, PubMed, the ACM Digital Library, Embase, Scopus, Ei Compendex, and APA PsycINFO. The articles identified through a search of these databases were screened using our eligibility criteria. The resulting set of screened articles was further augmented by additional relevant articles identified via citation mining. Three reviewers analyzed, discussed, and conducted data extraction on the final set of articles. Out of a total of 641 studies retrieved from the searched databases and subsequent citation mining, ten studies satisfied the inclusion criteria for this review. 
% Our analysis did not establish 

%To analyze this nascent field of study we conducted a scoping review; thereby, we analyzed the outcomes of the studies conducted so far in this field and found them limited due to a singular focus on the older adult population, use of small sample sizes, limited intervention durations, and other compounding factors. Nevertheless, the reviewed studies reported several positive outcomes, highlighting the potential social robots hold in this field.




\maketitle

\section{Introduction}\label{sec1}

% the state of sleep health crisis
Sleep is essential for maintaining good physical, emotional, and cognitive health.
% lack of sleep in adults
Despite sleep duration recommendations ranging from seven to nine hours per night for adults, 35.2\% of adults in the United States---and more than 42\% of single parents and factory workers---report getting less than seven hours of sleep nightly \cite{short_sleep}.
In adults, poor sleeping habits are associated with an increased likelihood of developing chronic diseases, increased risk of car accidents, lower psychomotor performance, and decreased productivity \cite{pub_health_crisis}. 
% the importance of sleep in children
For children, the reported sleep data is even more concerning, with 57.8\% of middle school and 72.7\% of high school students achieving less than the recommended nine to eleven hours of sleep per night \cite{short_sleep}; lack of proper sleep negatively impacts children's performance in school, mood regulation, cognitive processes, and general health \cite{dahl, wolfson, fredriksen, gruber}. 
Sleep disorders are quickly becoming recognized as a public health crisis, costing approximately \$400 billion annually in the US alone due to their growing prevalence \cite{sleep_matters}. 

% current technological solutions to the poor sleep health problem
Given the widespread sleep health crisis and the current shortage of sleep health professionals available to address it on an individual level \cite{losing_sleep}, technology-based interventions have become an important alternative in reducing barriers to and improving sleep health treatments. 
Two systematic reviews have evaluated the use of wearable and mobile consumer technologies in sleep health interventions \cite{mhealth_survey, wearable_survey}; these reviews highlight the potential of utilizing wearables (e.g., sleep trackers or smart watches) and mobile phones for sleep health monitoring and treatment \cite{wearable_survey}.
For example, mobile-phone-driven interventions in both auxiliary and alternative capacities have been reported to provide better sleep health outcomes relative to traditional treatments (e.g., cognitive behavioral therapy for insomnia) \cite{mhealth_survey}. Recent studies have also explored using virtual reality (VR) to improve users' sleep and have found similarly promising results: regular VR-based exercises have improved sleep quality in older adults and hospital patients \cite{older_vr_sleep, leuk_vr_sleep}, while VR-driven breathing exercises and meditations have positively affected sleep efficiency in adolescents and intensive care unit patients, respectively \cite{young_vr_sleep, icu_vr_sleep}. 
In all of these technology-based interventions, a key prerequisite to delivering sustained, individualized benefits is continuous user engagement. 

Unlike wearables, mobile phones, and VR, \textit{social robots} \cite{sar_foundation} possess unique qualities that allow them to motivate people to adhere to interventions beyond simple nudges (e.g., pop-up notifications). Robots have physical embodiments and are designed to engage with people through situated, multimodal social interactions; these qualities are vital in establishing social connections, sustaining long-term engagement, and offering personalized experiences, which are all important for successful behavioral interventions.
% Social robots have shown exciting potential to be used as a healthcare tool to augment individuals' health and well-being. 
Previous literature reviews have suggested the feasibility of using social robots in therapeutic interventions to obtain health outcomes such as reduced anxiety, pain, and stress and improved quality of life \cite{sar_review_children, sar_review_pyschosocial, sar_review_wellbeing}. 
In particular, social robots have been demonstrated to successfully aid in behavioral and cognitive therapies through social and emotional support mechanisms \cite{sar_foundation};
for example, social robots have been employed to improve mood and decrease stress, agitation, anxiety, and medication use in older adults with dementia \cite{sar_dementia, sar_dementia_rct} and have shown promise in delivering psycho-therapeutic treatments toward helping users achieve health goals \cite{sar_psychotherapeutic, scassellati2018improving}.

Despite compelling evidence exhibiting the effectiveness and potential of social robots in aiding behavioral interventions, their applicability in facilitating sleep health diagnoses and interventions has only been explored to a very limited degree.
To drive future research in this domain, we must critically examine the outcomes of existing studies on the use of social robots in sleep health interventions and identify gaps in present research; a recent systematic review and meta-analysis analyzed the impact of four different randomized control trials involving robots on adults' total sleep time however found no significant impacts \cite{store2022effect}. Given the nascent nature of this field, a less-rigid scoping review can help better establish the scope of current sleep health interventions using social robots and future opportunities.

In this paper, we critically evaluate relevant published experiments and findings to help inform the future development of social robots designed to aid in sleep interventions and discuss potential directions for prospective research. 

\section{Methods}\label{sec3}
We conducted a scoping review of available scientific literature following the methodology proposed by Peters et al. \cite{scoping_review_guidelines}; scoping reviews aim to summarize and disseminate research findings, identify research gaps, and make recommendations for future research \cite{daudt}. With a current lack of substantial literature exploring the use of social robots in the context of sleep health interventions, we elected to use the scoping review method as it is particularly effective in mapping germinal research with the goal of guiding future work in that area. 

\subsection{Search Strategy}\label{subsec2}
We defined our search string after an initial, limited review of the titles and abstracts of relevant literature to identify keywords and phrases. We initially built our search string with the following terms: ``social robots'' and ``sleep''. We further refined our search string and selected relevant databases in consultation with two academic librarians---one with an expertise in computer science and engineering literature and the other with an expertise in medical literature; the selected databases included IEEE Xplore, PubMed, the ACM Digital Library, Embase, Scopus, Ei Compendex, and APA PsycINFO.

We utilized the following search string to retrieve relevant articles from the aforementioned databases: ((companion OR assistive OR social OR service OR caregiver) AND (robot OR robot*)) AND (((sleep OR sleep*) OR ((``sleep initiation and maintenance disorder'' OR (sleep AND initiation AND maintenance AND disorders) OR insomnia)) OR (apnea OR apnoea))).

\begin{figure}[h!]
  \includegraphics[width=\textwidth]{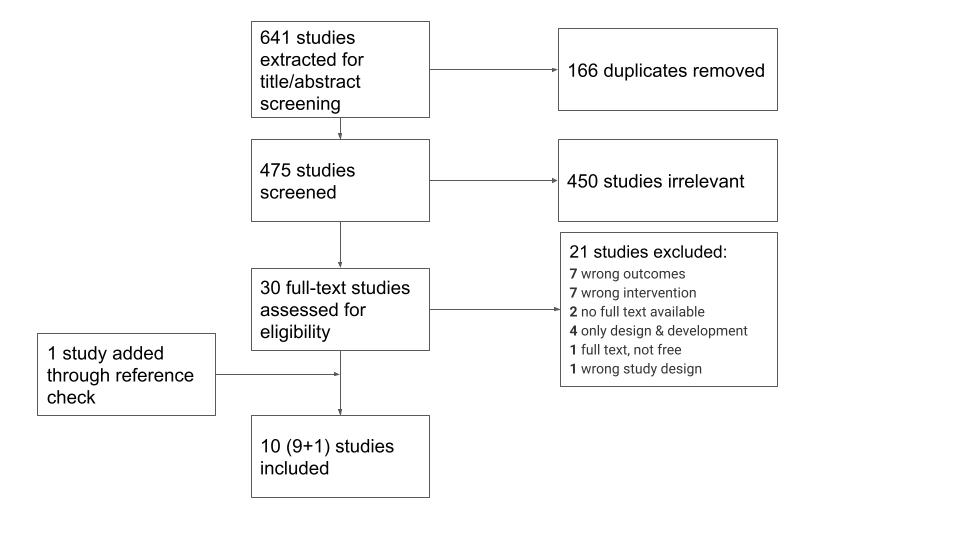}
  \caption{Preferred Reporting Items for Systematic Reviews and Meta-Analyses (PRISMA) diagram.}
  \label{fig:prisma}
\end{figure}

The search of the selected databases on March 9th, 2023 using the above search string yielded 641 results, which were then uploaded to Covidence, a systematic review software that helps organize and screen articles. The details of our study screening process are described in Fig \ref{fig:prisma}. 

\subsection{Selection of Relevant Publications}\label{subsec2}
Two research team members screened the titles and abstracts of our search results based on the following inclusion and exclusion criteria:\\
%JP: The numbered items should be indented more than the bolded "Inclusion criteria" and "Exclusion criteria" subheaders.
\vspace{3mm}
\textbf{Inclusion criteria}: 
\begin{enumerate}
    \item The study is published in a peer-reviewed journal or conference proceeding. 
    \item The study is written in English.
    \item A full-text version of the study is available.
    \item The publication assesses aspects of sleep health outcomes using social robots.
\end{enumerate}

\textbf{Exclusion criteria}: 
\begin{enumerate}
    \item The study does not report on sleep health interventions or diagnoses. 
    \item The publication solely describes robot development. 
    \item The publication focuses on a surgical intervention. 
    \item The publication is a review of existing literature.
    \item The study solely describes sleep intervention protocol or recruitment strategies.
\end{enumerate}

The articles remaining after this screening were retrieved for a full-text review; the same two team members read and discussed the appropriateness of each article to determine their inclusion in this scoping review. Any disagreements were resolved via further discussion of the criteria until a consensus was reached. 
The research team additionally conducted a reference check among the publications selected for full-text review to identify further articles of relevance. 
We note that no publication time limits were included in the above criteria, as the concept of ``social robots'' has only emerged in recent years.

\subsection{Data Extraction}\label{subsec2}

For each of the publications that passed the full-text screening, we extracted their 1) study characteristics, 2) research methodologies, and 3) research findings. 
Study characteristics include basic information such as the authors' names, the countries in which the studies were conducted, year of publication, and purpose of the study, while research methodologies focus on study design, setting, sample, intervention, duration, and outcome measurements. 
Finally, we summarized the main findings of the studies that aligned with the purpose of our review.

Full-text data extractions were conducted by three researchers---one with a background in engineering and robotics, the other two with backgrounds in health science. Each article underwent data extraction performed by one of the three researchers; the extraction was then double-checked by another member of the research team to ensure that the extracted contents were accurate. The team met weekly to discuss their progress and resolve any disagreements regarding data extraction via mutual consensus.

\subsection{Quality Appraisal}\label{subsec2}
Each publication was independently appraised by two of the three researchers using the Joanna Briggs Institute critical appraisal tools \cite{jbi}; specifically, the Checklists for Randomized Controlled Trials and Quasi-Experimental Studies were used to determine study quality. The three research team members then met and discussed to reach an agreement on the final quality appraisal. Please refer to Tables \ref{table:rct_asmt} and \ref{table:quasi_asmt} in the Supporting Materials section for our quality appraisal outcomes.

Additionally, we used the ROBIS tool to assess the risk of bias in our scoping review \cite{robis}. Three of this review's authors collectively answered the signalling questions and resolved disagreements through discussion and mutual consensus. The tool ultimately yielded a low risk of bias in our assessment. Table \ref{table:robis} in the Supporting Materials section presents the outcome of our ROBIS assessment and the rationales for any potential concerns.

\section{Results}\label{sec2}

An overview of key study characteristics of the final ten selected articles is provided in Table \ref{table:1}. We first detail the key attributes of the social robots used and the design of the interventions assessed in the included studies before concluding this section with the interventions' outcomes and limitations.

\begin{table}
    \caption{Summary of the reviewed studies.}
\input{tables/review_summary}
\label{table:1}
\end{table}

\subsection{Robots Used}\label{subsec2}
\begin{figure}[h!]
  \includegraphics[width=\textwidth]{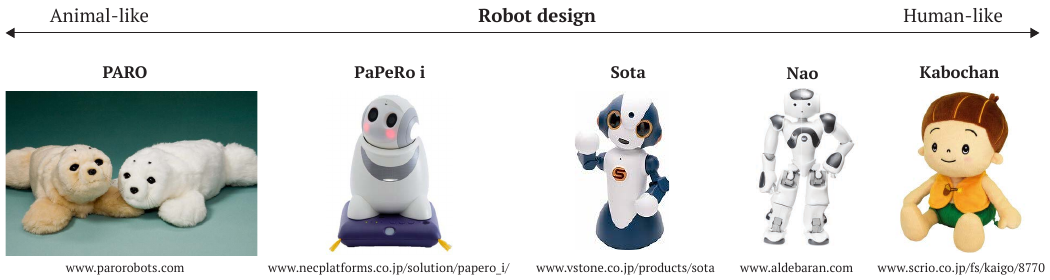}
  \caption{Social robots used in the reviewed studies.}
  \label{fig:sar}
\end{figure}

Comparing the social robots used in the reviewed studies helps to identify attributes that are critical in effectively driving sleep health interventions.
A total of six different social robots were used in the ten reviewed studies (Figure \ref{fig:sar}). 
Four studies used PARO, a robotic baby harp seal \cite{paro_website};
one study \cite{mizuno} used the PaPeRo (``Partner-type-Personal-Robot'') robot \cite{papero}; 
two studies \cite{obayashi, oda} used a humanoid robot called Sota \cite{sota}; and one study \cite{tanaka} employed the Kabochan Nodding Communication Robot \cite{kabochan}.
All of the aforementioned social robots were developed by Japanese firms and/or academic institutions. Nao, developed by Aldebaran Robotics, was another humanoid robot used in a study \cite{van2022interactive}.
Finally, one study \cite{peri} included in this review did not provide a detailed enough description of the robot it used.

\subsection{Physical Attributes}\label{subsec2}
Despite primarily executing tasks through social rather than physical interactions, a social robot's physical embodiment greatly impacts its performance and how others perceive it
\cite{sar_foundation}. 
Social robots are designed to faciliate social and empathic interactions, yet their physical attributes may differ greatly in terms of degree of human likeness---yet the shared physical characteristics (e.g., adorable eyes, pleasing shape, etc.) between the animal-like PARO, the childlike Kabochan, and the more robotic Sota, Nao and PaPeRo serve as evidence of the significance of a robot's embodiment in driving positive interactions.
PARO resembles a baby harp seal and is designed to be cute and calming with a soft, white, synthetic fur exterior \cite{paro_website}. Kabochan sports a similarly soft exterior while resembling a 3-year-old boy in form, voice, and movement \cite{kabochan}. PaPeRo, Nao and Sota, on the other hand, have hard-shell bodies but are equally adorable in shape. All of these robots also have charming eyes to further their friendly demeanor; the eyes of PaPeRo, Nao and Sota additionally feature embedded LED lights to augment their social interactions. These robots' pleasant, non-threatening embodiments help to drive positive interactions, while their animal-like or non-realistic humanoid designs temper user expectations; this is critical given the sensitive application domain and current limited capabilities of even state-of-the-art artificial intelligence \cite{sar_embodiment}.

\begin{table}
\caption{Overview of the reviewed studies.}
\centering
\input{tables/robot_overview}
\label{table:2}
\end{table}

\subsection{Sensors}\label{subsec2}
Sensors help robots perceive the world and are therefore crucial in both powering social interactions and effectively driving interventions. 
% vision and sound
The robots used in the evaluated studies were all equipped with cameras or light sensors and microphones, allowing them to ``see,'' ``hear,'' and respond appropriately. 
% touch / haptics
PARO, Nao, and Kabochan also have an array of tactile sensors installed across their bodies to detect and react to users' touches. 
% temperature 
PARO has additional temperature sensors, similar to PaPeRo's temperature and humidity sensors; although it is unclear whether these capabilities were utilized in the reviewed experiments, these environmental sensors hold the potential to modify a social robot's intervention based on the state of the physical environment. 
% smart sensors
Sota may also be linked with smart sensors (e.g., smart watches), thus allowing researchers to potentially personalize interactions and interventions based on factors such as heart rate and body temperature.
Overall, the sensors available to a social robot are fundamental in facilitating dynamic interactions and intervention execution.

\subsection{Social Interaction Capabilities}\label{subsec2}
The user experience provided by a social robot is a critical factor in the effectiveness of its interventions and directly affects its ability to drive engaging social interactions. A common social interaction feature across the robots in the reviewed studies was face tracking, which in some cases also allowed for face recognition. Face tracking allows for the simulation of eye contact, which significantly improves interaction quality by communicating attentiveness and indicating interpersonal interest \cite{davidhizar_1992}. Furthermore, all of the robots additionally expressed themselves to users via the sound modality, with PARO being limited to animal-like sounds and the more humanoid robots like PaPeRo using natural-language-based speech interactions with varying degrees of autonomy. 

% robot motions
Social robots also tailor their physical motions relative to their sensor inputs to produce lively, friendly interactions. The goal of personalizing a robot's interactions to its user's actions is the development of a positive relationship and consequent enhancement of the intervention experience.
PARO leverages its tactile sensors to power interaction through movement in its tail, flippers, and eyes during petting; it also responds to sounds, including its name and any words its user frequently repeats. 
Similarly, Kabochan can verbally respond---as well as sing and nod---in response to its user's touch and spoken words. 
PaPeRo interacts by blinking the LEDs on its cheeks, mouth, and ears to express rich emotions; it also executes face tracking in response to human words. 
Similarly, Sota interacts using a cloud-based speech dialogue system and is able to track and remember specific faces. Nao's interactions are similarly dynamic, powered by its speech dialogue and two 2D camera systems.

\subsection{Sleep Sensors}\label{subsec2}
% on-body sensors
Social robots are equipped with sensors designed primarily to enable social interactions; therefore in order to track and/or support users' sleep health, additional external sensors were used in the reviewed studies. See Table \ref{table:sleep_vars} for a summary of the sleep variables evaluated and the sensors used to measure them.

Wrist- and arm-based actigraphy sensors were the most widely utilized, with four of the nine included studies relying on them to collect sleep statistics including sleep time, sleep efficiency, and wake after sleep onset (WASO) by utilizing accelerometer data. Despite being designed as less intrusive relative to electroencephalography (EEG) sensors, these on-body sensors were reported to not have been well-tolerated by certain users \cite{thodberg, moyle}. Furthermore, actigraphy methods have been criticized for overestimating sleep, thus undermining their reliability---especially for individuals with chronic conditions \cite{conley_actigraphy}.

% off-body sensors
Given the sensitive user groups these investigations have targeted so far, some studies also elected to use completely off-body sensors. For instance, a series of infrared sensors was deployed in participants' living environments to collect their wake-up times, bedtimes, and sleep duration \cite{mizuno}. 
Similarly, sheet-shaped body vibrometers (SBV) allow for constant, noninvasive monitoring of care recipients' sleep patterns while in bed \cite{obayashi}; in one reviewed study, an SBV informed a social robot about the nighttime awakenings of study participants in a nursing home, allowing the robot to intervene when necessary \cite{obayashi}. Although SBVs and infrared sensor networks are plausible off-body alternatives for collecting sleep data, they require considerable effort to deploy and are not yet commercially available like on-body actigraphy sensors are.

Three studies did not use any sleep sensors at all, instead relying on observations from care providers to estimate subjects' sleep habits \cite{peri} or utilizing questionnaires to collect data on sleep outcomes, such as nocturnal sleeping hours and difficulty in initiating sleep \cite{tanaka, van2022interactive}. Although these survey-based methods do not provide the same degree of fidelity and granularity in their measurements relative to on- and off-body sensors, they do provide insights into the more user-centered and subjective aspects of sleep health.

\begin{table}
\caption{Summary of the sleep variables evaluated and sleep sensors used in the reviewed studies.}
\centering
\input{tables/sleep_vars_overview}
\label{table:sleep_vars}
\end{table}

\subsection{Robot-Driven Intervention and Study Design}\label{subsec2}
In this section, we describe the design details of the reviewed robot-driven interventions to provide key insights toward understanding their effectiveness and interpreting their results.
All of the studies included in this review measured the change in certain sleep variables as a result of their designed interventions either as a primary or secondary focus.
Tables \ref{table:1}, \ref{table:2}, and \ref{table:sleep_vars} summarize the key intervention details of the reviewed studies.
Below, we first present the results of the interventions and then discuss their limitations.

\subsubsection{Intervention Overview}\label{subsubsec2}
Five studies evaluated the impact of psychosocial interactions on sleep parameters, three studies explored the use of information support models to improve sleep health outcomes, one study measured the impact of social-robot-driven interactive wake-ups to minimize sleep inertia in participants and a study evaluated the impact of a robot-facilitated sleep hygiene education.

% psycho-social interactions
One exploratory study investigated the impact of individual, facilitated tactile and verbal interactions with pets or pet-like aids (i.e., dogs, PARO, plush toys) on sleep patterns (i.e., sleep duration, sleep efficiency, sleep fragmentation) and the psychiatric well-being of older adults \cite{thodberg}. 
Similarly, the effect of individual but non-facilitated PARO-based interventions on participants' sleep---as measured by time spent lying down, awake, and in sleep (light, deep, and very deep sleep)---was evaluated in other studies \cite{pu, moyle}; researchers attempted to leverage tactilely and verbally stimulating interactions with PARO to reduce pain and agitation while inducing a calming effect on participants, hypothesizing a carryover effect on nighttime sleep quality.
Another study investigated changes in the sleep patterns (i.e., sleep efficiency, WASO, nocturnal awakenings longer than five minutes, total sleep time) of older adults as a result of facilitated, PARO-assisted group activities, speculating that an increase in social and tactile stimulation would lead to better sleep \cite{joranson}. 
In a different approach than the aforementioned PARO-based experiments, the Kabochan robot was deployed to participants' homes to develop a friendly user relationship through affective behavior, such as nodding while talking \cite{tanaka}; the impact of this intervention on participants' cognitive functions was measured along with the following sleep outcomes: nocturnal sleeping hours, difficulty in initiating sleep, difficulty in maintaining sleep, and early morning awakening. The investigators claimed that companionship with the robot resulted in lower stress levels (as measured by cortisol in saliva) and cited previous work suggesting that lower stress levels lead to improved sleep quality when rationalizing their results \cite{tanaka}.

% information support model
In contrast, three of the retrieved studies used an information support model wherein the social robot delivered useful information---such as reminders to take medication at appropriate times---as part of their intervention.
In one experiment, social robots were stationed in two low-level care lounges at retirement homes, where they interacted with people and were incorporated into regular activities to assess their impact on daytime sleepiness \cite{peri}; these robots were expected to be more socially stimulating than the existing televisions in the lounges, thus leading to fewer participants sleeping during the day.
In another study, PaPeRo was installed in the houses of participants living alone to support their independent living by providing them with important reminders (such as when to take out the garbage) with the aim of introducing a regular schedule in their lives and thereby stabilizing their sleep patterns \cite{mizuno}; the impact of PaPeRo's information support intervention on the subjects' wake-up times, bedtimes, sleep duration, number of sensor firings at night, and other measures of daily activity was recorded.
Likewise, Sota robots were installed in a care home to assist residents during nighttime wake-ups and to alert support staff; in case of a nighttime wake-up, the robot interacted with the participant to keep them in bed and reduce their chance of falling while support staff were en route \cite{obayashi}. To evaluate this intervention, total sleep time, total time in bed (TTIB), sleep latency time, sleep efficiency, WASO, total number of times leaving bed after sleep onset, mean respiratory rate during TTIB, and mean heart rate during TTIB were measured using SBVs. 

% minimize sleep inertia
Sota was also used to execute an intervention designed to minimize sleep inertia---i.e., the period of reduced alertness and impaired cognition after waking up \cite{sleep_inertia, oda}; Sota robots were installed in the homes of college students to interactively wake them up while minimizing their sleep inertia. To appraise the efficacy of this intervention, sleepiness after wake-up was measured using the Stanford Sleepiness Scale \cite{stanford_sleep_scale}.

%sleep hygiene education
Taking a different approach to providing useful information towards improving sleep health, a study used the Nao robot to facilitate an interactive sleep hygiene education to improve sleep hygiene in children receiving anti-cancer treatment at a pediatric oncology outpatient clinic \cite{van2022interactive}. Nao interacted with the participants to gain an understanding of their sleep habits, co-created improved sleep routines, and built repertoire through personalized questions and engaging activities such as dancing together. The impact of this approach was evaluated using a Dutch language version of the Children’s Sleep Hygiene Scale \cite{harsh2002measure}. 

It is worth noting that none of the reviewed studies reported any simultaneous or parallel human-driven or pharmacological sleep interventions along with the studied social-robot-driven intervention.

\subsubsection{Intervention Location, Duration, and Frequency}\label{subsubsec2}
Six of the reviewed studies were conducted in nursing homes or residential care facilities and one study was done in a pediatric oncology outpatient clinic whereas the remaining three studies were administered in subjects' homes. The more controlled environments of the nursing homes, outpatient facilities and care facilities simplify the deployment challenge while allowing access to the most vulnerable groups; however, despite their complexities, home-based interventions facilitate the exploration of how social robots may help augment subjects' independent living, which is a valued factor for many individuals, especially in older age groups \cite{reframing_aging}.

Apart from an intervention's location, its duration is also critical in clinically appraising its efficacy; the reviewed studies had intervention durations ranging from two days to twelve weeks, with an average of 6.9 weeks. The frequency of an intervention alongside its duration also provides valuable context for interpreting its results. The frequency of the PARO-based interventions ranged between thirty minutes daily to ten minutes biweekly. The shortest intervention duration was two days, which makes the reported positive outcomes difficult to use as scientific evidence, especially given the novelty effect in using new technologies such as social robots \cite{oda}. Certain interventions were limited to only nighttime or daytime operations due to their target tasks \cite{obayashi, peri}. Two studies did not mention turning the robot off at any point, so they very well may have been on for all twenty-four hours of the day \cite{mizuno, tanaka}; robotic intervention that is available around-the-clock may have potential benefits, such as increased trust and reliance---however, it may also have unintended consequences, such as occasionally interrupting users' sleep \cite{obayashi}. The sleep hygiene education intervention, an outlier in terms of duration, featured a one-off 10-minute session.

\subsubsection{Intervention Target Population and Sample Size}\label{subsubsec2}
As could be expected from the prevalence of nursing homes and care facilities as intervention locations, older individuals were by far the most common target population of the reviewed studies. Only two of the reviewed investigations did not target older adults, instead choosing college students and pediatric oncology patients as their target population \cite{oda, van2022interactive}. All of the PARO-based studies targeted more vulnerable subgroups of older adults, such as elderly patients living with dementia, Alzheimer's, or chronic pain; very complex sleep health issues manifest due to these medical conditions and exploration of how robots can be used in this context may help inform design guidelines for general robotic sleep health interventions.

There was a wide range of sample sizes for the user studies in the reviewed articles, ranging from fourteen to 280. The PARO-based studies tended to have larger sample sizes since the same robot could be used for multiple participants. Similarly, deploying robots in shared spaces of care facilities allowed for larger sample sizes as a single robot was able to conduct interventions for several participants \cite{peri}. Conversely, the home-based interventions were restricted to smaller sample sizes as each subject needed their own robot for a longer period of time; this restriction may explain why short intervention durations occurred in sync with relatively higher sample sizes \cite{oda}. Overall, we observed that the reviewed sample sizes had a strong correlation with the frequency and duration of the investigated interventions. 

Social robots are relatively expensive to build and require multidisciplinary expertise to deploy; these barriers have a strong impact on both the duration and frequency of social-robot-driven interventions and their sample sizes. Researchers have had to make compromises to balance these study parameters, consequently limiting the nature of scientific inquiry in this field.

\subsection{Sleep Outcomes}\label{subsec2}
The reviewed studies did not clearly establish how or to what degree robot-driven interventions impact sleep health; however, they did provide several promising and statistically significant indicators for certain key sleep parameters, highlighting the potential for social robots' future in this domain. Table \ref{table:sleep_vars} details the sleep parameters investigated by the reviewed studies and indicates whether a statistically significant result was reported for each parameter.

\subsubsection{Sleep Outcomes}\label{subsubsec2}
There were several indicators that robot-driven interventions led to significant improvements in overall sleep duration; nocturnal sleeping hours and total sleep time tended to increase in social robot intervention groups \cite{tanaka, joranson}. There was also evidence signaling a positive impact on sleep duration: one PARO intervention group had a greater increase in their nightly sleep period \cite{pu}. Interestingly, one study found that sleep duration increased in one of the data collection weeks only in the animal therapy control condition---rather than the PARO robot therapy condition \cite{thodberg}; this finding emphasizes the differences between robotic and real animals while also suggesting that better emulation of key characteristics of animal-based therapies (such as affective tactile responses) may yield better outcomes in robot-based interventions. 

An important aspect of sleep duration is a stable sleep-wake rhythm, which is often disturbed in individuals living alone, thus affecting their quality of sleep and life \cite{beibei_sleepwake}. Notably, an information support robot intervention stabilized the sleep-wake schedule of older adults living alone by reducing their sleep duration and inducing faster wake-up times. These results validated the experiment's hypothesis that people following regular schedules tend to be more active ``morning types.'' \cite{mizuno, morningness}.

\subsubsection{Sleep Efficiency}\label{subsubsec2}
Sleep disturbances are critical to the degradation of overall sleep quality. An increase in sleep efficiency (the percentage of time in bed and actually asleep), a reduced number of nocturnal awakenings, and reduced awakenings after sleep onset were reported for one social robot intervention group \cite{joranson}. Similarly, a study featuring a communication robot observed a decrease in difficulty in maintaining sleep \cite{tanaka}.

\subsubsection{Daytime Sleepiness}\label{subsubsec2}
The presence of social robots decreased the amount of daytime sleepiness and the proportion of people sleeping in low-care nursing home lounges during the day; however, these findings must be qualified given the large variance in the individual results of participants \cite{peri}. Correspondingly, a greater increase in daytime wakefulness and a greater reduction in daytime sleep for a robot intervention group (as compared to a control group) was reported \cite{pu} and an increase in the amount of daytime activity was recorded for one robot-intervention condition \cite{mizuno}.

These outcomes have important implications, as daytime sleepiness is associated with an increased risk of several common health conditions, such as cardiovascular mortality, cognitive deficits, depression, disrupted nighttime sleep, and higher risk of falling \cite{daytime_sleep_impacts, martin2006daytime}. For context, 80\% of adverse fall-related incidents occur during the night for hospitalized patients \cite{nighttime_falls}; this statistic further emphasizes the value of improving overall nighttime sleep quality, which includes age-appropriate sleep duration, better sleep efficiency, and decreased daytime sleepiness.

\subsubsection{Sleep Inertia}\label{subsubsec2}
Sleep inertia---the period of reduced alertness and impaired cognition after waking up \cite{sleep_inertia}---is a condition largely overlooked by society even as it affects large proportions of adolescents and younger adults \cite{sleep_inertia_impact_young} with serious health ramifications; for instance, sleep inertia impairs cognitive performance with an equivalence to forty hours of sleep deprivation \cite{sleep_inertia_current_insights}. 
Robot-driven interactive wake-ups led to less sleepiness in the morning relative to a non-interactive alarm clock condition \cite{oda}; while this result indicates that social robots may help mitigate sleep inertia, the reporting study had a limited sample size (n=22) and intervention duration (only two days).

\subsubsection{Sleep Hygiene}\label{subsubsec2}
Sleep hygiene refers to behavioral practices, such as calming pre-sleep routines and stable bedtimes, that may influence sleep initiation and maintenance \cite{durand1998sleep, harsh2002measure}. Robot-driven sleep hygiene education was received positively by children and their parents and led to statistically significant improvements in self-reported sleep hygiene scores. Despite their survey-based data collection limiting the validity of their reported outcomes, the fact that a simple 10-minute one-off interaction with a social robot potentially leads to improved sleep habits highlights the value of multi-modal engaging interactions for sleep health education.

Though empirical evidence has suggested social robots' potential in improving certain aspects of sleep quality, the effects of robot-driven sleep interventions are mixed. One study did not reveal any statistically significant sleep outcomes for care recipients at a nursing home, but reported that caregivers felt empowered by their new ability to track care recipients' sleep efficiency and physical conditions \cite{obayashi}; this investigation also noted that the robots' occasional interruptions of participants' sleep and the short duration of the study may have led to the absence of significant quantitative findings. Likewise, another investigation documented that a PARO-based intervention positively affected participants' motor activity but had no effect on any measured sleep patterns; however, it is worth noting that reduced nighttime motor activity has a positive association with reduced nighttime falls \cite{moyle}.

\subsection{Study Barriers and Limitations}\label{subsec2}
Most of the reviewed studies acknowledged the limitations of their experiments to a certain extent; limitations reported across the studies included a limited number of participants \cite{tanaka, mizuno, joranson}, restricted participant demographics \cite{oda, mizuno}, inadequate intervention duration \cite{tanaka, thodberg}, lack of sustained impact post-intervention \cite{pu}, uncertainty in collected data \cite{pu, moyle, joranson,van2022interactive}, insufficient analysis of sleep data \cite{mizuno}, inadequacies in robot interactions \cite{peri, oda, obayashi, van2022interactive}, the absence of a randomized control trial design \cite{peri}, and high risk of the novelty effect \cite{joranson}. It should be noted that the shortcomings reported above are common across many studies in this review and not only the ones wherein they were explicitly acknowledged.

The most commonly \textit{unacknowledged} limitations were inadequate descriptions of the finer details of the explored robot-driven interventions and minimal discussion on the nature of the interactions between the subjects and the robots, rendering it very difficult to fully contextualize the interventions and the robots' roles within them and thus complicating the assessment of the robot's true impact on its user's sleep health. On a similar note, several studies failed to provide detailed descriptions of the sleep scales they used or only used investigator-developed sleep measures; the absence of such key details, along with a lack of standardized sleep measures, hinders a proper comparison of the results of the different studies. It should also be noted that long-term post-intervention effects were not studied by the reviewed publications in any meaningful manner. There was also a general lack of acknowledgement or discussion on the potential privacy issues that arise when deploying digital agents in personal living spaces. Moreover, the studies were limited to only Japanese, Northern European, and Oceanic contexts, which may pose an issue as cultural implications may affect human-robot interactions and thus the efficacy of any of these robot-driven sleep health interventions.

\section{Discussion}\label{sec12}

This review has highlighted a wide array of robot-aided sleep health outcomes while also elucidating a broad range of intervention designs and contexts; our analysis strongly indicates that using social robots in support of sleep health has the potential for a positive impact despite a lack of current strong clinical evidence. The reviewed studies reported improved outcomes for sleep efficiency, total sleep time, nocturnal awakenings, and sleep inertia. However, the studies' limitations introduce uncertainty in establishing causality between sleep health and the interventions themselves; the heterogeneity of the robots deployed, a focus on largely older target populations, the intervention designs themselves, and a lack of standardization of measurements only add to this uncertainty. Despite the largely exploratory nature of the work in this field so far, there are strong indications that building upon these existing social-robot-driven sleep health interventions and further embracing the unique strengths of social robots (e.g., motivational interactions, companionship, etc.) may lead to much stronger, clinically significant outcomes. Based on our analysis, we provide further discussion below on future work to improve sleep health via social-robot-driven interventions. 

\subsection{Robust Study Design}\label{subsec2}
As previously mentioned, many of the analyzed studies were limited to small sample sizes and short intervention durations, adding uncertainty to their reported results and additionally impeding the development of a robust understanding of the longer-term impacts of social-robot-driven interventions on improving sleep health. Long-term randomized controlled trials should be attempted to neutralize the effects of factors such as the novelty effect and to help establish clinical results. 

A major factor contributing to the small sample sizes in studies involving social robots is the cost of these robots in general---on average, the social robots used in the reviewed studies cost around \$6,000 USD each. Furthermore, commercially available social robots are not very customizable for research purposes; on the other hand, developing an in-house social robot may reduce costs and allow it to be tailored to the requirements of a given experiment---but also requires a significant amount of financial investment, engineering work, and human expertise. Cost concerns drastically reduce the number of research groups that are able to conduct large-scale studies involving social robots; therefore there is a critical need for a flexible yet affordable social robot platform to drive further research in this area.

Our analysis also indicated that the social robots used in the reviewed studies were not specifically designed for sleep health interventions; currently, researchers must adapt their interventions to the capabilities of social robots designed for general social interactions, a fact reflected in the general designs of the reviewed interventions. The social robots presently available for use are viable for exploratory studies, but as this field matures and targets more specific sleep mechanisms, the functional requirements for these social robots will shift. Either the robots must be designed more specifically and with requirements stemming from a deeper understanding of the sleep health field or they must be modular or customizable enough to be easily converted to meet such requirements. The additional research requirements expected as this field matures further underscore the need for a flexible yet affordable social robot platform to foster more robust research.

\subsection{Comprehensive Study Documentation}\label{subsec2}
The reviewed studies largely provided limited descriptions of the roles the robots played within their interventions and failed to sufficiently characterize the nature of the robots' interactions with their human users, severely inhibiting accurate analysis and assessment of the robots' actual impact on experimental outcomes. This deters the establishment of any clinical relevance of social-robot-driven sleep health interventions and hinders further work in this field as other researchers do not have a precise or definitive foundation to build upon. Future studies should elaborate on the details of their robots' intervention roles and the nature of their interactions to allow for a more accurate appraisal of their impact on any reported outcomes.

On a similar note, the reviewed studies failed to report the individual sleep health backgrounds (i.e., existing sleep disturbances, self-reported poor sleep quality, insomnia, etc.) of their participants. They also did not report any information on participants' relevant personal histories, such as their typical diet, exercise regimen, physical activity level, medication use, or alternative interventions previously or concurrently tried. Although most studies elaborated on the sleep health challenges their targeted sub-populations are generally known to face, the absence of individual participants' sleep and personal histories deprives the reported outcomes of valuable context; future research should collect and present these crucial data-points to underscore their results.

Finally, most of the reviewed studies only provided shallow justification for their intervention design choices; specifically, there was a lack of robust grounding of intervention plans and details in any previous work---especially within any sleep science literature. This is somewhat acceptable given the exploratory nature of the studies conducted so far; however, future work should attempt to rationalize intervention designs with relevant references to sleep science research in order to validate their experimental hypotheses and improve study outcomes.

\subsection{Fundamental Challenges in Human-Robot Interaction}\label{subsec2}
A key justification for deploying social robots in the healthcare domain is their potential to boost user motivation to maximize compliance with interventions through effective interactions. Good communication is critical for maintaining motivation \cite{comm4motivation} and is a fundamental aspect of effective human-robot interaction (HRI). Good communication becomes even more potent in the context of long-term deployments where the novelty effect eventually wanes. Negative effects of sub-optimal human-robot interactions were reported in the reviewed studies as a result of a robot's interactivity being incompatible with a user's physical capabilities \cite{peri}; a lack of smooth, rich, context-appropriate conversations \cite{oda, obayashi}; and discomforting physical appearance \cite{obayashi}. It is interesting to note that with the aid of an expert observer, interactions with social robots tended to be more positive \cite{peri}; thus, to maximize the positive impact of a social robot's deployment---especially in a sensitive health care domain like sleep health---researchers should consider user experience of past experiments and attenuate prior shortcomings for their studies. Furthermore, there is a need for better sensing and communication models to foster more engaging, situation-suitable interactions with users in the long term. Additionally, the ability to understand a user's preferences regarding the physical embodiment of their robotic companion and the personalization of a robot's appearance in accordance with said preferences may further enhance user experience.

Another fundamental factor that may impact the efficacy of an intervention is how and where a social robot engages with its users, which in turn depends on the robot's mobility; for instance, one of the reviewed studies detailed how mobile robots in nursing homes may increase opportunities for user interaction \cite{peri}, while another study reported on the development of a smart, mobile alarm clock to counter oversleeping and sleep fragmentation \cite{bedrunner}. Empowering social robots with more robust mobility may open new avenues for their application and efficiency; however, socially aware, safe navigation is a nuanced and challenging problem \cite{gao2021evaluation} which requires further research before deployment in the real world. Two recent studies reported mixed results of insomnia interventions using Somnox, a non-social sleep robot with haptic interactions \cite{store2022effects, store2022technically}; exploring haptic interactions for social robots for sleep may be an interesting avenue for future research.

\subsection{Intervention Type and Targets}\label{subsec2}
Only five studies from this review sample cited sleep health outcomes as the primary focus of their interventions \cite{oda, joranson, moyle, obayashi, van2022interactive}. The remaining studies reported positive sleep health outcomes, but had originally designed general interventions without a focus on sleep; these interventions generally did not leverage prior work in sleep medicine to inform their design. Despite the lack of a scientific focus on sleep health, the resulting positive outcomes are encouraging; it may be that leveraging past research in sleep science to design and drive future interventions will help yield even better outcomes.

There is a large focus on older adults in this research area, with only one of the reviewed studies focusing on a younger age group. There is valid justification for such a strong focus, as large portions of older adults, especially ones with chronic conditions, tend to experience overall poorer sleep quality; for instance, 71\% of people with dementia have sleep disorders such as insomnia, daytime sleepiness, restless leg syndrome, REM sleep behavior disorder, etc. \cite{dementia_sleep}. Still, different age groups and those with certain medical conditions have different sleep requirements and challenges, demanding further work toward understanding how social robots may facilitate the sleep needs of younger adults and children. There is also a lack of research on treating specific sleep disorders (e.g., snoring, bruxism, restless leg syndrome) which impact more than fifty million people in the US alone \cite{sleep_disorders}. Using social robots to drive health interventions for a wider population may help improve sleep interventions generally---for instance, by driving the discovery of better models for motivating intervention adherence and collecting larger data sets to help classify specific sleep disorders.

Lastly, none of the reviewed studies attempted to diagnose any sleep disorders through the use of social robots. Having captured a wide range of sleep data by deploying these robots in people's living spaces, there is a lost opportunity to leverage such data to detect and characterize sleep disorders and customize interventions around them.

\subsection{Sleep Health Measures}\label{subsec2}
Despite being the gold standard for non-invasive on-body sleep health measurements, arm- and wrist-based actigraphy sensors are cited to be unreliable in their recordings and have been criticized for requiring long wear time to ensure the validity of their data \cite{accelerometers, sensewear_validity}; furthermore, it was reported that these sleep trackers may not be suitable for use with elderly individuals \cite{moyle}. On the other hand, deploying a completely non-invasive network of infrared sensors for the collection of sleep statistics requires a significant installation effort and may not be suitable for many common living situations \cite{mizuno}. A sleep measure that has been increasingly used in sleep research is the Sleep Profiler, an in-home, three-channel (EEG, electromyography [EMG], and electrooculogram [EOG]) sleep monitor that allows for detailed sleep assessments beyond the basic sleep measures collected via actigraphy; this system facilitates the collection of deeper information regarding sleep spindles, REM latency, and micro-arousals---all data which help contextualize an individual's sleep health---yet still suffers from the same usability shortcomings as other actigraphy methods.

Therefore, there is an urgent need for a cheap, comprehensive, easily deployable, non-invasive, off-body sleep tracking system. Recent work has explored using radio frequency (RF) signals for non-intrusive sleep state detection and other health measurements \cite{rf_sleep, rf_heart}; further research into improving RF-signal-based health statistic measurements may lead to the development of the accessible sensors required for social robots to drive more dynamic and personalized interventions---especially for the most vulnerable and sensitive populations. 

\subsection{Ethics}\label{subsec2}
There are several ethical dilemmas that must be addressed when deploying social robots to drive health interventions. For vulnerable populations such as older adults, researchers must account for the emotional bond that may form between a deployed system and its human user and quantify the degree of trauma that may be caused by their separation at the end of the study. Additionally, there is a growing danger of inducing Turing deceptions in participants as social robots become increasingly more competent at social interactions \cite{SAR_ethics}. There are also several privacy issues with deploying robots in the home health care context that should be addressed promptly \cite{SAR_ethics}, as users' privacy concerns may hinder their trust in and reliance on intervention systems and thus dampen the efficacy of those interventions. There must also be further discussion on the role of social robots in sleep health interventions; for example, whether social robots should independently drive interventions or if they should facilitate interventions in collaboration with human experts is a critical question yet to be addressed.

\subsection{Limitations of the Current Review}\label{subsec2}

Our review has its own limitations. First, our search was limited to articles published in English; given the relatively large proportion of work in this domain originating from Japan, we may have missed certain articles of relevance published in different languages. Moreover, our search only included papers that were published during or before March 2023; thus we may have missed articles of relevance that have been published since. It should also be noted that this review was not registered as registration was not required for a scoping review following the methodology by Peters et al. \cite{scoping_review_guidelines}.

\section{Conclusion}\label{sec13}

This scoping review reveals that social robots have the potential to improve specific sleep health outcomes; however, this field of inquiry is still in its nascence---limited due to a largely unitary focus on older adults, a lack of robust grounding of intervention design in sleep health science, and small sample sizes and short intervention durations in conducted user studies. The presence of positive study results despite the wide array of limitations indicates the promise that social robots hold in this field and should motivate further work on using social robots to improve sleep health outcomes.

\section{Supplementary Materials}\label{supp_mat}
%\section{Quality Assessment}\label{secROBIS}

\begin{table}[h]
\caption{ROBIS Assessment}
\centering
\input{tables/robis_table}
\label{table:robis}
\end{table}

\bigskip

\begin{table}[h]
\caption{Quality Assessment of Reviewed Randomized Control Trials}
\centering
\input{tables/rct_qual}
\label{table:rct_asmt}
\end{table}

\bigskip

\begin{table}[h]
\caption{Quality Assessment of Reviewed Quasi-Experimental Trials}
\centering
\input{tables/quasi_qual}
\label{table:quasi_asmt}
\end{table}

\clearpage

\backmatter

\bmhead{Acknowledgments}

This work was partially supported by the Malone Center for Engineering in Healthcare at the Johns Hopkins University. The authors would like to thank Jaimie Patterson for proofreading this paper.

\section*{Declarations}

The authors declare no competing interests. The authors confirm that the data supporting the findings of this study are available within the article and its Supplementary Material. Any further data analysis information is available from the corresponding author by request.

%\begin{appendices}




%\end{appendices}


\bibliography{sn-bibliography}% common bib file

\end{document}

%% file: tables/review_summary.tex
    \begin{tabular}{|>{\hspace{0pt}}m{0.169\linewidth}|>{\hspace{0pt}}m{0.142\linewidth}|>{\hspace{0pt}}m{0.088\linewidth}|>{\hspace{0pt}}m{0.098\linewidth}|>{\hspace{0pt}}m{0.223\linewidth}|>{\hspace{0pt}}m{0.212\linewidth}|} 
\hline
\textbf{Study (Year)} & \textbf{Study Design} & \textbf{Sample Size} & \textbf{Age} & \textbf{Research Setting} & \textbf{Participant Type} \\ 
\hline
Joranson et al. (2021) \cite{joranson} & Randomized control trial & 60 & 62–95 (range) & Nursing home & Older adults w/ dementia \\ 
\hline
Mizuno et al. (2021) \cite{mizuno} & Quasi-experimental trial & 14 & 82.8 (mean) & Self-support facilities  participants' homes & Older adults \\ 
\hline
Pu et al. (2020) \cite{pu} & Randomized control trial & 43 & 65–97 (range) & Nursing home & Older adults w/ dementia  chronic pain \\ 
\hline
Obayashi et al. (2020) \cite{obayashi} & Quasi-experimental trial & 25 & 85.9 (mean) & Nursing home & Older adults \\ 
\hline
Oda et al. (2020 \cite{oda} & Quasi-experimental trial & 22 & 20–24 (range) & Participants' homes & College students \\ 
\hline
Moyle et al. (2018) \cite{moyle} & Randomized control trial & 175 & 84 (mean) & Nursing home & Older adults w/ dementia \\ 
\hline
Thodberg et al. (2016) \cite{thodberg} & Randomized control trial & 100 & 79–90 (range) & Nursing home & Older adults \\ 
\hline
Peri et al. (2016) \cite{peri} & Quasi-experimental trial & \par{}N/A & N/A & Nursing home & Older adults \\ 
\hline
Tanaka et al. (2012) \cite{tanaka} & Randomized control trial & 34 & 66–84 (range) & Participants' homes & Older adults (women only) \\ 
\hline
van Bindsbergen et al. (2022) \cite{van2022interactive} & Quasi-experimental trial & 28 & 8–12 (range) & Pediatric oncology outpatient clinic & Children \\
\hline
\end{tabular}

%% file: tables/robot_overview.tex
\begin{tabular}{|>{\hspace{0pt}}m{0.095\linewidth}|>{\hspace{0pt}}m{0.05\linewidth}|>{\hspace{0pt}}m{0.4\linewidth}|>{\hspace{0pt}}m{0.3\linewidth}|} 
\hline
\textbf{Robot} & \textbf{Ref} & \textbf{Robot Intervention} & \textbf{Intervention Frequency/Duration} \\ 
\hline
PARO & \cite{joranson} & facilitated, group activity centered around interactions with robot & 30-minutes session twice per week for 12 weeks \\ 
\hline
~ & \cite{moyle} & individual, non-facilitated pet-like interaction with robot & 15-minute session 3 times per week for 10 weeks \\ 
\hline
~ & \cite{thodberg} & individual, facilitated tactile and verbal  interaction with robot, dog, or soft toy & 10-minute session twice per week for 6 weeks \\ 
\hline
~ & \cite{pu} & individual, non-facilitated pet-like interaction with robot & 30-minute session 5 days per week for 6 weeks \\ 
\hline
PaPeRo & \cite{mizuno} & informative interaction to support independent living (e.g., reminders on when to wake up, go to bed, eat, take medication, go out, watch TV, take out the garbage, etc.) & 4 weeks, continuous \\ 
\hline
Kabochan & \cite{tanaka} & establishment of friendly relationship through affective behaviors & 8 weeks, continuous \\ 
\hline
Sota & \cite{obayashi} & detection and reporting of and assistance with sleep interruptions & 4 weeks, continuous \\ 
\hline
~ & \cite{oda} & interactive conversation as a means of waking user & 2 days, continuous \\ 
\hline
NAO & \cite{van2022interactive} & interactive sleep hygiene education program & 10 minute session, one-off \\ 
\hline
Unknown & \cite{peri} & user interaction and entertainment, digital communication facilitation, vital sign tracking, and incorporation into various additional activities & 12 weeks, continuous \\
\hline
\end{tabular}

%% file: tables/sleep_vars_overview.tex
\begin{tabular}{|>{\hspace{0pt}}m{0.3\linewidth}|>{\hspace{0pt}}m{0.281\linewidth}|>{\hspace{0pt}}m{0.065\linewidth}|>{\hspace{0pt}}m{0.271\linewidth}|} 
\hline
\textbf{Construct Investigated} & \textbf{Data Collection Tool} & \textbf{Studies} & \textbf{Statistically Significant Effect? (Control/s)} \\ 
\hline
\textbf{\textit{Sleep-Duration-Related Outcomes}} &  &  &  \\ 
\hline
Sleep duration~ & Questionnaire & \cite{oda} & Not applicable \\ 
\hline
 & Infrared sensor & \cite{mizuno} & Yes \\ 
\hline
Total time in bed & Sheet-shaped body vibrometer (SBV)~ & \cite{obayashi} & No \\ 
\hline
Time spent lying down & Actigraphy (SenseWear 8.0 Activity Armband)~ & \cite{pu} & No \\ 
\hline
 & Actigraphy (SenseWear 8.0 Activity Armband)~ & \cite{moyle} & No \\ 
\hline
Sleep time (light sleep, deep sleep, very deep sleep)~ & Actigraphy (SenseWear 8.0 Activity Armband)~ & \cite{pu} & Yes \\ 
\hline
 & Actigraphy (SenseWear 8.0 Activity Armband)~ & \cite{moyle} & No \\ 
\hline
Nocturnal sleep hours & Questionnaire & \cite{tanaka} & Unclear \\ 
\hline
Actual sleep time & Actiwatch 4 (CamNTech) & \cite{thodberg} & No \\ 
\hline
Total sleep time (TST). & Sleep actigraph (ActiSleep+) & \cite{joranson} & Yes \\ 
\hline
\textbf{\textit{Sleep-Fragmentation-Related Outcomes}} &  &  &  \\ 
\hline
\par{}Sleep efficiency~ & Sleep actigraph (ActiSleep+) &  \cite{joranson} & Yes \\ 
\hline
 & Sheet-shaped body vibrometer (SBV)~ & \cite{obayashi} & No \\ 
\hline
 & Actiwatch 4 (CamNTech) & \cite{thodberg} & No \\ 
\hline
WASO (Wake after sleep onset) & Sleep actigraph (ActiSleep+) & \cite{joranson}  & Yes \\ 
\hline
 & Sheet-shaped body vibrometer (SBV)~ & \cite{obayashi} & No \\ 
\hline
Fragmentation index & Actiwatch 4 (CamNTech) & \cite{thodberg} & No \\ 
\hline
Difficulty in maintaining sleep & 4-level scale & \cite{tanaka} &  \\ 
\hline
Total time leaving bed after sleep onset & Sheet-shaped body vibrometer (SBV)~ & \cite{obayashi} & No \\ 
\hline
Number of awakenings longer than 5 minutes (NA  5) & Sleep actigraph (ActiSleep+) & \cite{joranson} & Yes \\ 
\hline
Time spent awake & Actigraphy (SenseWear 8.0 Activity Armband)~ & \cite{pu} & Yes \\ 
\hline
 & Actigraphy (SenseWear 8.0 Activity Armband)~ & \cite{moyle} & No \\ 
\hline
\textbf{\textit{Sleep-Inititation-Related Outcomes}} &  &  &  \\ 
\hline
Bedtime & Infrared sensor & \cite{mizuno} & No \\ 
\hline
Difficulty initiating sleep~ & 4-level scale & \cite{tanaka} & Unclear \\ 
\hline
Sleep latency~ & Sheet-shaped body vibrometer (SBV)~ & \cite{obayashi} & No \\ 
\hline

\textbf{\textit{Wake-Up-Related Outcomes}} &  &  &  \\ 
\hline
Wake-up time~ & Infrared sensor & \cite{mizuno} & Yes \\ 
\hline
Early morning awakening & 4-level scale & \cite{tanaka}  & Unclear \\ 
\hline
Sleepiness (after wake-up) & Stanford Sleepiness Scale & \cite{oda}  & Yes \\ 
\hline
\textbf{\textit{Sleep-Hygiene-Related Outcomes}} &  &  &  \\ 
\hline
Sleep Hygiene Score~ & 6-level scale & \cite{van2022interactive} & Yes \\
\hline
\end{tabular}

%% file: tables/robis_table.tex
\begin{tabular}{|>{\hspace{0pt}}m{0.2\linewidth}|>{\hspace{0pt}}m{0.1\linewidth}|>{\hspace{0pt}}m{0.6\linewidth}|} 
\hline
\textbf{Domain} & \textbf{Concern} & \textbf{Rationale for concern} \\ 
\hline
Concerns regarding specification of study eligibility criteria & Low risk & Eligibility criteria were clear and unambiguous. All signaling questions were answered "Yes" or "Probably Yes". The eligibility criteria were restrict to English language studies as a result of the reviewers' language limitations, potentially introducing minor publication bias. We targeted future development and focused on the robots' design details, using full-text when available. \\ 
\hline
Concerns regarding methods used to identify and/or select studies & Low risk & All signaling questions were answered "Yes”. The process for both screening titles and abstract assessment of full text papers was reported and included multiple reviewers. \\ 
\hline
Concerns regarding used to collect data and appraise studies & Low risk & All signaling questions were answered "Yes”. All articles were assessed independently by a minimum of two reviewers and the appropriate data were abstracted independently. Study quality was formally assessed using an appropriate tool. \\ 
\hline
Concerns regarding the synthesis & Low risk & All signaling questions were answered "Yes”. We analyzed all sleep outcomes and robot designs. Furthermore, we addressed the limitations of every study and used the ROBIS tool to measure our bias. \\ 
\hline
\textbf{Risk of bias in the review} & \textbf{Rating} & ~ \\ 
\hline
Risk of bias & Low risk & The above assessment shows no concerns with our review process. The potential limitations of these studies are fully addressed in the Discussion section. This review's conclusions reflect its results appropriately.\\
\hline
\end{tabular}

%% file: tables/rct_qual.tex
\begin{tabular}{|>{\hspace{0pt}}m{0.5\linewidth}|>{\hspace{0pt}}m{0.069\linewidth}|>{\hspace{0pt}}m{0.063\linewidth}|>{\hspace{0pt}}m{0.071\linewidth}|>{\hspace{0pt}}m{0.05\linewidth}|>{\hspace{0pt}}m{0.063\linewidth}|} 
\hline
Checklist Items & Joranson et al. \cite{joranson} & Moyle et al. \cite{moyle}& Thodberg et al. \cite{thodberg}& Pu et al. \cite{pu}& Tanaka et al. \cite{tanaka} \\ 
\hline
1.~ Was true randomization used for assignment of participants to treatment groups? & 1 & 1 & 0 & 1 & 0 \\
2.~ Was allocation to treatment groups concealed? & 1 & 1 & 0 & 0 & 0 \\
3.~ Were treatment groups similar at the baseline? & 1 & 1 & 0 & 1 & 1 \\
4.~ Were participants blind to treatment assignment? & 0 & 1 & 0 & 0 & 0 \\
5.~ Were those delivering treatment blind to treatment assignment? & 0 & 0 & 0 & 0 & 0 \\
6.~ Were outcomes assessors blind to treatment assignment? & 1 & 1 & 0 & 1 & 0 \\
7.~ Were treatment groups treated identically other than the intervention of interest? & 1 & 1 & 1 & 1 & 1 \\
8.~ Was follow-up complete and if not, were differences between groups in terms of their follow up adequately described and analyzed? & 1 & 1 & 1 & 1 & 1 \\
9.~ Were participants analyzed in the groups to which they were randomized? & 1 & 1 & 1 & 1 & 1 \\
10.Were outcomes measured in the same way for treatment groups? & 1 & 1 & 1 & 1 & 1 \\
11.Were outcomes measured in a reliable way? & 1 & 1 & 1 & 1 & 1 \\
12.Was appropriate statistical analysis used? & 1 & 1 & 1 & 1 & 1 \\
13.Was the trial design appropriate and were any deviations from the standard RCT design (individual randomization, parallel groups) accounted for in the conduct and analysis of the trial? & 1 & 1 & 1 & 1 & 1 \\ 
\hline
Total score (out of 13) & 11 & 12 & 7 & 10 & 8 \\
\hline
\end{tabular}

%% file: tables/quasi_qual.tex
\begin{tabular}{|>{\hspace{0pt}}m{0.546\linewidth}|>{\hspace{0pt}}m{0.075\linewidth}|>{\hspace{0pt}}m{0.06\linewidth}|>{\hspace{0pt}}m{0.083\linewidth}|>{\hspace{0pt}}m{0.063\linewidth}|>{\hspace{0pt}}m{0.108\linewidth}|} 
\hline
Checklist Items &  Mizuno et al. \cite{mizuno} & Peri et al. \cite{peri}& Obayashi et al. \cite{obayashi}& Oda et al. \cite{oda} & van Bindsbergen et al. \cite{van2022interactive}\\  
\hline
1.~ Is it clear in the study what is the “cause” and “effect” (i.e. there was no confusion about which variable came first)? & 1 & 1 & 1 & 1 & 1 \\
2.~ Were the participants included in any comparisons, similar? & 1 & 0 & 0 & 0 & 0 \\
3.~ Were the participants included in any comparisons receiving similar treatment/care, other than the exposure or intervention of interest? & 1 & 1 & 1 & 1 & 1 \\
4.~ Was there a control group? & 0 & 1 & 0 & 0 & 0 \\
5.~ Were there multiple measurements of the outcome both pre- and post-intervention/exposure? & 1 & 0 & 1 & 1 & 0 \\
6.~ Was follow-up complete and if not, were differences between groups in terms of their follow-up adequately described and analyzed? & 1 & 0 & 1 & 0 & 1 \\
7.~ Were the outcomes of participants included in any comparisons measured in the same way? & 1 & 1 & 1 & 1 & 1 \\
8.~ Were outcomes measured in a reliable way? & 0 & 0 & 1 & 1 & 1 \\
9.~ Was appropriate statistical analysis used? & 1 & 1 & 1 & 1 & 1 \\ 
\hline
Total score (out of 9) & 7 & 5 & 7 & 6 & 6 \\
\hline
\end{tabular}